\documentclass[dvipsnames]{article} %
\usepackage{colm2024_conference}

\usepackage{booktabs}
\usepackage{enumitem}
\usepackage{wrapfig}
\usepackage{algorithm}
\usepackage{algpseudocode}
\usepackage{graphicx}
\usepackage{subfigure}
\usepackage[misc]{ifsym}

\usepackage{microtype}
\usepackage{amsmath}
\usepackage{colortbl}
\usepackage[utf8]{inputenc}
\usepackage[T1]{fontenc}
\definecolor{lightgray}{rgb}{0.9,0.9,0.9}
\usepackage{caption}
\usepackage{subcaption}
\usepackage{setspace}
\usepackage{url}
\usepackage{multirow}
\usepackage{tabularx}
\usepackage{blindtext}
\usepackage{pgfplots}
\pgfplotsset{compat=1.18} 
\usepackage{tikz}
\usetikzlibrary{er,positioning,bayesnet}
\usepackage{makecell}
\usepackage{tipa}
\usepackage{siunitx}
\usepackage{nicefrac}
\usepackage{listings}
\usepackage[raster,skins, most]{tcolorbox} %
\usepackage{xltabular}
\usepackage{adjustbox}
\usepackage{xurl}
\usepackage{rotating}
\usepackage[normalem]{ulem}

\usepackage{fontawesome}

\useunder{\uline}{\ul}{}

\usepackage{amsmath,amsfonts,bm}

\def\eqref#1{equation~\ref{#1}}

\def\1{\bm{1}}

\DeclareMathAlphabet{\mathsfit}{\encodingdefault}{\sfdefault}{m}{sl}
\SetMathAlphabet{\mathsfit}{bold}{\encodingdefault}{\sfdefault}{bx}{n}

\renewcommand{\ghlink}{https://github.com/Alibaba-NLP/DeepResearch}

\usepackage{makecell}
\usetikzlibrary{tikzmark}
\makeatletter
\newcommand*\myfontsize{%
  \@setfontsize\myfontsize{7}{8}%
}
\makeatother

\usepackage{fontawesome}   

\definecolor{uclablue}{RGB}{159, 195, 224}

\definecolor{uclagold}{RGB}{255, 240, 180}

\definecolor{aliceblue}{RGB}{255, 238, 241}

\definecolor{cadmiumgreen}{rgb}{0.0, 0.42, 0.24}

\definecolor{myred}{rgb}{0.7, 0.3, 0.0}
\definecolor{myblue}{rgb}{0.2, 0.3, 0.6}
\definecolor{babygreen}{rgb}{0.85, 0.97, 0.85}

\definecolor{purple1}{RGB}{126, 107, 196}
\definecolor{purple2}{RGB}{199, 158, 207}
\definecolor{purple3}{RGB}{214, 200, 255}
\definecolor{purple4}{RGB}{254, 240, 255}

\definecolor{deepblue}{RGB}{48, 58, 82}

\makeatletter

\newcommand{\symboletongyi}{\raisebox{0pt}{~\includegraphics[scale=0.012]{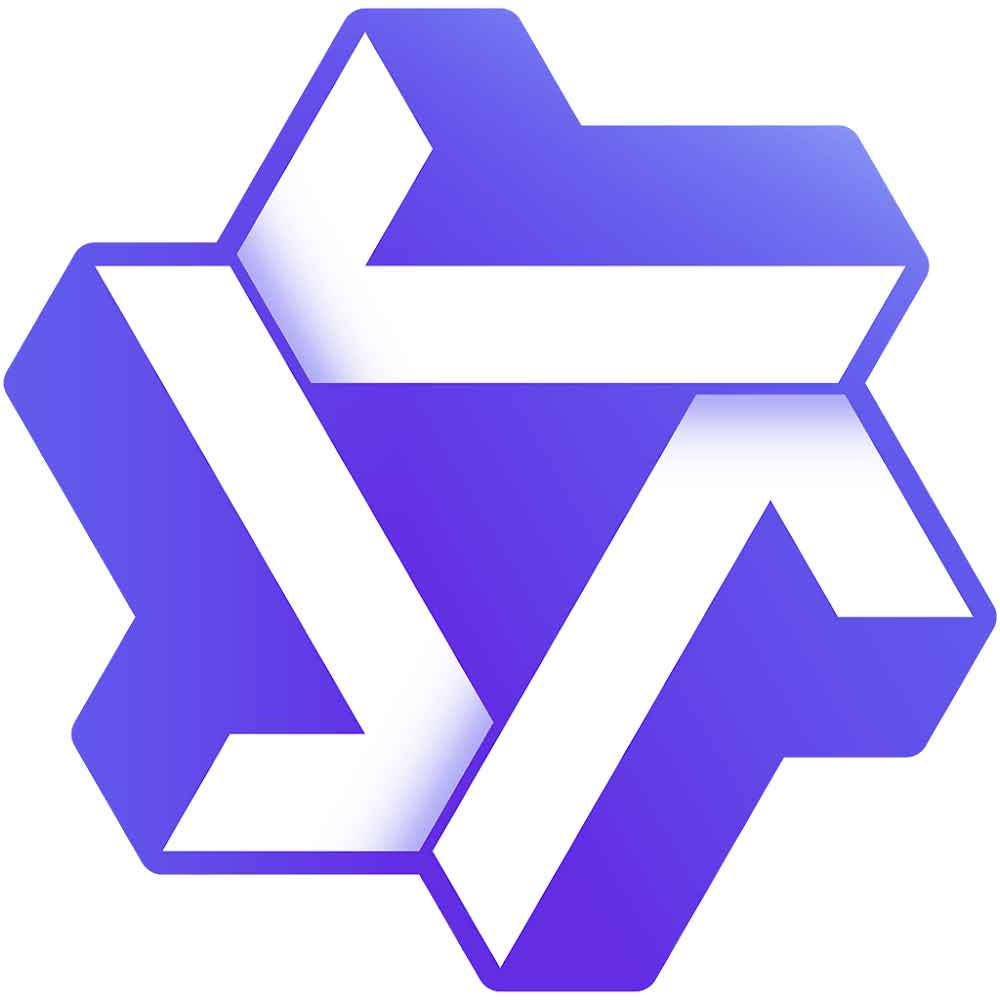}}~}

\definecolor{deepPurple}{HTML}{330066}

\definecolor{uclablue_old}{rgb}{0.15, 0.45, 0.68}
\hypersetup{
    breaklinks,
    citecolor=uclablue_old,
    colorlinks=true,
}

\newtcolorbox{mybox}[2][]
  {colback = black!5!white, colframe = black!75!black, fonttitle = \bfseries,
    colbacktitle = black!100!black, enhanced, before upper={\fontsize{8}{11}\obeyspaces\obeylines\selectfont}, fontupper=\selectfont,
    attach boxed title to top left={yshift=-2.2mm,xshift=4mm},
    title=#2,#1}

\newcommand{\equal}{\textsuperscript{\dag}}

\definecolor{darkgreen}{rgb}{0.0, 0.5, 0.0}

\author{%
\small{Baixuan Li\equal$^{(\textrm{\Letter})}$, Jialong Wu\equal, Wenbiao Yin\equal$^{(\textrm{\Letter})}$, Kuan Li, Zhongwang Zhang, Huifeng Yin, Zhengwei Tao, Liwen Zhang, Pengjun Xie, Jingren Zhou, Yong Jiang$^{(\textrm{\Letter})}$}%
  \\[1em]               
  {\fontsize{10pt}{11pt}\selectfont          
Tongyi Lab\symboletongyi, Alibaba Group}\\
}

\usepackage{caption}
\usepackage{subcaption}
\usepackage{amssymb}

\usepackage{xcolor}

\usepackage{pifont}
\newcommand{\cmark}{\textcolor{green!65!black}{\ding{51}}}
\newcommand{\xmark}{\textcolor{red!75!black}{\ding{55}}}

\definecolor{mydarkgray}{gray}{0.2} 

\begin{document}

\title{Nested Browser-Use Learning for Agentic Information Seeking}

\maketitle

\begingroup
\renewcommand\thefootnote{\equal}
\footnotetext{Equal contribution.}
\endgroup

\begingroup
\renewcommand\thefootnote{$^{\textrm{\Letter}}$}
\footnotetext{Correspondence to: \texttt{baixuan@seu.edu.cn}, \texttt{\{yinwenbiao.ywb, yongjiang.jy\}@alibaba-inc.com}.}
\endgroup

\begin{abstract}
Information-seeking (IS) agents have achieved strong performance across a range of wide and deep search tasks, yet their tool use remains largely restricted to API-level snippet retrieval and URL-based page fetching, limiting access to the richer information available through real browsing. While full browser interaction could unlock deeper capabilities, its fine-grained control and verbose page content returns introduce substantial complexity for ReAct-style function-calling agents. To bridge this gap, we propose Nested Browser-Use Learning (\textbf{NestBrowse}), which introduces a minimal and complete browser-action framework that decouples interaction control from page exploration through a nested structure. This design simplifies agentic reasoning while enabling effective deep-web information acquisition. 
Empirical results on challenging deep IS benchmarks demonstrate that NestBrowse offers clear benefits in practice.
Further in‑depth analyses underscore its efficiency and flexibility.
\end{abstract}
\section{Introduction}

\begin{wrapfigure}{r}{0.5\linewidth}
    \centering
    \includegraphics[width=1\linewidth]{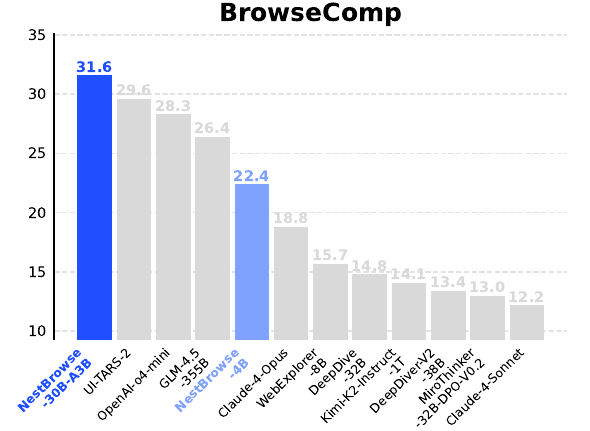}
    \caption{Performance of the proposed NestBrowse on the challenging deep IS benchmark BrowseComp.}
    \label{fig:first_fig}
\end{wrapfigure}

Information-seeking (IS) agents \citep{tongyidr, grokdr, skywork2025deepresearch, kimi_researcher, geminidr},  powered by large language models (LLMs) have achieved strong performance on a variety of challenging wide and deep search tasks \citep{mialon2023gaia, bc_en, bc_zh, xbench, wong2025widesearch}. These agents typically follow a ReAct-style function-calling paradigm \citep{yao2023react}, interleaving reasoning with tool invocation to incrementally gather and process evidence. 
While IS agents can, in principle, leverage a wide range of external information sources, most existing systems primarily rely on the web, which remains the most comprehensive repository of publicly accessible knowledge\footnote{Unless otherwise specified, IS agents in this work refer to text-only, ReAct-style LLM-driven systems that use the web as their information source to gather evidence.}.

Most existing IS agents \citep{wu2025webdancerautonomousinformationseeking, li2025websailornavigatingsuperhumanreasoning, tao2025webshaper, zhao2025repurposing, tao2025webleaper, li2025parallelmuse} model IS with two tools: \texttt{search}, which retrieves query-related URLs, titles, and snippets, and a goal-conditioned \texttt{visit}, which fetches the Markdown content of a given URL. While effective for shallow IS, this abstraction provides only partial access to the web. In practice, substantial information and functionality are exposed only through browser-level interactions or state changes, including client-side rendering, dynamic content loading, form-mediated workflows, multi-step navigation, and page-internal online functionalities that go beyond static content retrieval. Such content is neither reliably surfaced by search engines nor recoverable via a single static URL fetch, rendering IS formulations based solely on \texttt{search} and \texttt{visit} inherently incomplete.

Recent work \citep{openaidr, kimi_researcher, wang2025ui} has explored equipping IS agents with real browser interactions. However, there is no widely adopted standard for modeling browser actions, and the inherent complexity of browser-use makes it challenging to abstract diverse interaction patterns into a tool interface suitable for IS agents \citep{yu2025browseragent, hu2025osagents}. Moreover, browser pages often contain large amounts of raw and highly redundant information. For ReAct-style IS agents, naively injecting all browser outputs into the context is suboptimal due to excessive redundancy \citep{kerboua2025focusagent, wu2025resum, fang2025memp}, and can be impractical when individual pages exceed feasible context limits. Consequently, how to design a simple yet complete browser toolkit, together with an effective interaction paradigm that enables IS agents to use such tools efficiently, remains an open challenge.

To address this challenge, we propose \textbf{Nested Browser-Use Learning (NestBrowse)}, which consists of three key components. (i) We design a minimal yet functionally complete browser toolkit that provides four core actions, \texttt{search}, \texttt{visit}, \texttt{click}, and \texttt{fill}, enabling comprehensive web information access with low tool complexity. (ii) Building on this toolkit, we introduce the \textit{Nested Browser-Use Framework} that decouples browser interaction into an outer loop of tool-integrated reasoning and an inner loop of intra-page exploration, enabling controlled and incremental information flow into the agent’s context under practical context constraints. (iii) Finally, we develop a multi-task imitation learning formulation that internalizes these nested browser-use capabilities into a single IS agent model by jointly training outer-loop reasoning and inner-loop evidence extraction.

Building on the proposed NestBrowse, we train two IS agent models at different scales, namely NestBrowse-4B and NestBrowse-30B-A3B. Across four highly challenging deep IS benchmarks in both English and Chinese, both models exhibit consistently strong performance. These results support a key insight for IS tasks: with appropriate browser tool abstractions and interaction strategies, small-scale agent models can acquire effective browser-use capabilities and successfully solve complex deep IS problems.

\section{Related Work}

\paragraph{Tool-Integrated Reasoning.}

Tool-integrated reasoning (TIR) equips models with external tools and enables them to interleave reasoning with tool invocation to solve complex tasks \citep{lin2025understanding, xue2025simpletir}. By extending capabilities beyond parametric knowledge, this paradigm supports effective problem solving in settings that require information access, or external computation.

Prior work has identified several common principles for effective TIR \citep{shen2024llm, liu2024toolace}, with tool interface design playing a central role \citep{patil2024gorilla}. In particular, toolkits are typically kept minimal and low in complexity to reduce decision burden \citep{shen2023hugginggpt, qin2024tool}, avoid confusion during reasoning, and maintain efficiency and smooth information flow between tool execution and subsequent reasoning \citep{deng2023mind2web, liuagentbench}.

Motivated by these principles, we observe that the diversity of low-level browser operations makes naive action aggregation impractical; thus, we propose a minimal yet functionally complete browser toolkit and the \textit{Nested Browser-Use Framework} for efficient agentic reasoning over compact, goal-aligned page content.

\paragraph{Deep Information Seeking.}

Deep information-seeking (IS) tasks \citep{webwalker, bc_en, xbench} require agents to acquire and integrate hard-to-find information from external sources through iterative reasoning and exploration. Unlike conventional multi-hop QA \citep{yang2018hotpotqa, ho2020constructingmultihopqadataset}, where explicit entities and intermediate clues are provided, deep IS tasks begin from vague or underspecified prompts and demand that agents actively infer latent clues, entities, and relations. This process can be viewed as the incremental construction of an implicit entity-relation graph \citep{li2025websailornavigatingsuperhumanreasoning, tao2025webshaper, li2025parallelmuse}, driven by hypothesis generation, evidence gathering, and verification. Such a detective-style reasoning process poses substantial challenges, even for humans.

Accordingly, IS agents interleave reasoning with information acquisition to gather task-relevant evidence. Prior work largely focuses on web-based IS agents using ReAct-style tool invocation \citep{webwalker, wu2025webdancerautonomousinformationseeking, qiao2025webresearcher, tongyidr}, where a key challenge remains the design of effective interaction abstractions and tool-use strategies for information acquisition.

In this work, we address these challenges by introducing a practical methodology that enables IS agents to leverage more realistic browser-based interactions, allowing them to acquire more comprehensive and better-structured web information for solving complex deep IS tasks.

\section{Nested Browser-Use Learning}

In this section, we first present a minimally complete browser toolkit (\S\ref{sec:toolkit}), followed by the nested browser-use framework that enables IS agents to efficiently and effectively leverage this browser toolkit for web information acquisition (\S\ref{sec:framework}). We then introduce a multi-task imitation learning paradigm to internalize IS and browser-use capabilities into a base model (\S\ref{sec:learning}).

\subsection{Minimally Complete Browser Toolkit} \label{sec:toolkit}

We implement a headless browser backend\footnote{Notably, our setup considers only textual page content and does not incorporate any visual information.} in \textit{Playwright} for programmatic web interaction, parsing each page from raw HTML into a \textbf{\textit{semantic DOM snapshot}} that exposes interactive-element identifiers for subsequent actions while presenting structured, LLM-readable content.

To abstract browser interactions into tools usable by IS agents, we partition web information into two categories, $\mathcal{I} = \mathcal{I}_{\text{static}} \cup \mathcal{I}_{\text{dynamic}}$: 
(i) \textit{static information} $\mathcal{I}_{\text{static}}$, accessible through a single page load without in-page interaction; and 
(ii) \textit{dynamic information} $\mathcal{I}_{\text{dynamic}}$, exposed only via browser-level interactions such as client-side rendering, incremental loading, or user-triggered actions. 
Mainstream IS tool abstractions focus on query-based retrieval (\texttt{search}) and page-level fetching (\texttt{visit}), which suffice for $\mathcal{I}_{\text{static}}$ but provide limited access to $\mathcal{I}_{\text{dynamic}}$. At the same time, overly fine-grained browser tool formulations are undesirable, as increased action-space complexity substantially amplifies the decision burden for IS agents \citep{xu2025cognitive}, hindering efficient information acquisition.

Accordingly, we balance functional completeness for web information access with \textbf{minimal} tool complexity, and propose a minimally complete browser toolkit consisting of four tools:
\begin{itemize}[itemsep=0pt, topsep=0pt]
    \item \texttt{search}: Performs batched Google queries and returns the top-10 ranked results for each.
    \item \texttt{visit}: Fetches webpage from a URL and extracts information relevant to the given goal.
    \item \texttt{click}: Interacts with a clickable element, potentially triggers a page transition, and extracts content relevant to the given goal.
    \item \texttt{fill}: Types text into form fields or other editable elements within the current page.
\end{itemize}
The \texttt{search} and \texttt{visit} follow the standard tool configurations adopted by most existing IS agents and serve as the primary means for accessing static information $\mathcal{I}_{\text{static}}$. To cover the full spectrum of dynamic information $\mathcal{I}_{\text{dynamic}}$, we introduce two additional browser actions, \texttt{click} and \texttt{fill}. Together, these four tools complete the information access pathway for web-based IS agents, ensuring functional completeness while introducing only two additional actions and keeping tool complexity within a highly usable range.

It is worth noting that operations such as scrolling and in-page search, which are commonly included in existing browser toolkits, are omitted in our design; the rationale is discussed in \S \ref{sec:framework}.

\subsection{Nested Browser-Use Framework} \label{sec:framework}

\begin{figure*}[htbp]
    \centering
    \includegraphics[width=1\textwidth]{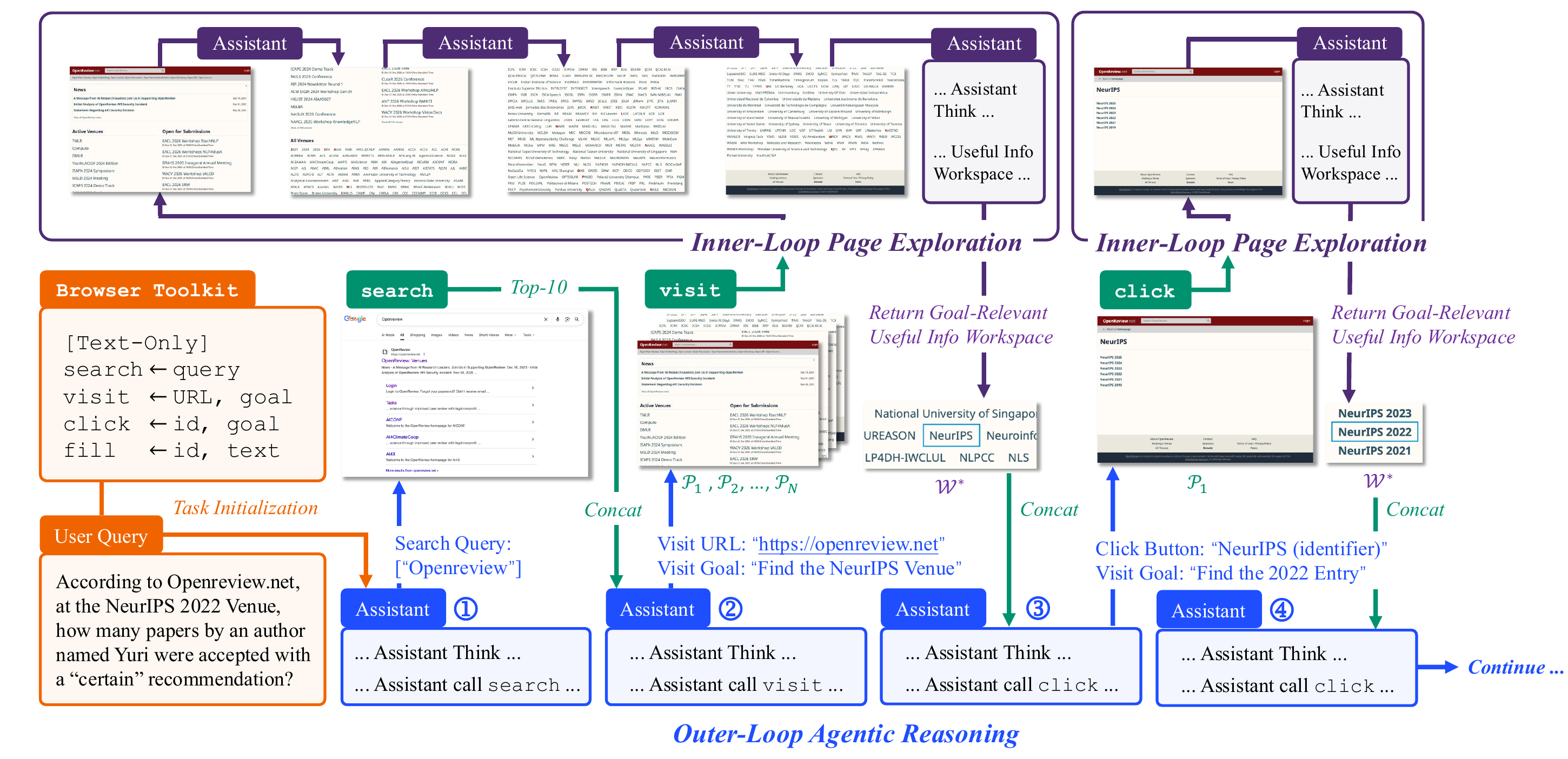}
    \caption{Overview of the Nested Browser-Use Framework. The outer loop interleaves reasoning and tool calls to solve the user task. Page-transition actions trigger an inner loop for intra-page exploration, which extracts and returns goal-relevant content to the outer loop, forming a nested interaction structure.}
    \label{fig:method}
\end{figure*}

\paragraph{Why Nested Browser-Use Matters.} 

Although the proposed browser toolkit enables functionally complete access to web information, raw page content is often highly redundant. A single page can easily exceed $64\mathrm{K}$ tokens and may even surpass $1\mathrm{M}$ tokens, rendering it infeasible for IS agents operating under typical context limits of $128\mathrm{K}$ or $256\mathrm{K}$ \citep{deepseekv3, yang2025qwen3}. Truncating overlong pages is a common workaround \citep{wu2025webdancerautonomousinformationseeking, li2025websailornavigatingsuperhumanreasoning}, but inevitably risks discarding information.

From a \textit{goal-driven} perspective, the objective of an IS agent is to acquire only the subset of page information necessary to resolve a task goal $g$. For a visited page with content $\mathcal{P}$, only a small subset $\mathcal{P}_g \subset \mathcal{P}$ is goal-relevant, while most content $\mathcal{P} \setminus \mathcal{P}_g$ is extraneous. 
Injecting full page content into the agent’s context wastes token budget and impairs subsequent reasoning. Accordingly, we \textbf{exclude} scrolling and in-page search, as they merely limit content exposure per page read without improving goal-directed information acquisition, resulting in inefficient browser-use for IS tasks.

\paragraph{How Nested Browser-Use Works.} To bridge the gap between browser-use and IS tasks, we decouple browser interaction into an outer loop of tool-integrated reasoning and an inner loop of goal-driven page exploration. This framework is termed \textit{nested} since the inner loop is fully contained within specific outer-loop tool invocation (Figure \ref{fig:method}).

The outer loop follows a standard ReAct-style function-calling paradigm. Let the agent context at outer-loop step $t$ be $c_t$, and let $\mathcal{T}$ denote the browser toolkit. The agent operates by
\begin{equation}\label{eq:outer_policy}
(a_t, \eta_t) \sim p_\theta(\cdot \mid c_t), \qquad a_t \in \mathcal{T},
\end{equation}
where $p_\theta$ denotes the parameterized IS agent model, $a_t$ is the selected tool at step $t$, and $\eta_t$ includes tool arguments. Executing the tool produces a response
\begin{equation}\label{eq:tool_response}
r_t = \mathrm{Exec}(a_t, \eta_t),
\end{equation}
which is appended to the agent’s context to form the agent state for the next reasoning step,
\begin{equation}\label{eq:context_update}
c_{t+1} = \mathrm{Update}(c_t, a_t, \eta_t, r_t).
\end{equation}
The outer loop repeats until a task-level termination condition is satisfied, including successful task completion or predefined resource constraints (e.g., sequence length or the number of tool calls).

When the invoked tool transitions the agent into a new page, an inner loop is instantiated for intra-page exploration.
Let $g_t$ denote the goal passed from the outer loop, and let $\mathcal{P}$ denote the raw textual content of the visited page. We partition $\mathcal{P}$ into a sequence of segments $\{\mathcal{P}_i\}_{i=1}^{N}$ and explore them incrementally under a goal-driven procedure. Concretely, the agent maintains a temporary workspace $\mathcal{W}$ initialized as $\emptyset$ and iteratively updates it as
\begin{equation}\label{eq:workspace}
\mathcal{W} \leftarrow \mathcal{W} \cup f(\mathcal{P}_i, g_t),
\end{equation}
where $f(\mathcal{P}_i, g_t)$ extracts information from segment $\mathcal{P}_i$ that is relevant to the goal $g_t$. Only goal-relevant content is accumulated in $\mathcal{W}$, while irrelevant segments are discarded. The inner loop terminates once the exploration of all the segmented page content is complete. 
The interaction between the outer and inner loops is formalized by defining tool execution via inner-loop aggregation. Specifically,
\begin{equation}\label{eq:nested_exec}
\mathrm{Exec}(a_t, \eta_t)
=
\begin{cases}
\mathcal{W}^\star(g_t, \mathcal{P}), & \text{if } a_t \in \mathcal{T}_{\text{page}}, \\
\mathrm{Exec}_{\mathrm{base}}(a_t, \eta_t), & \text{otherwise},
\end{cases}
\end{equation}
where $\mathcal{W}^\star(g_t, \mathcal{P})$ denotes the workspace produced upon termination of the inner loop under goal $g_t$ on page $\mathcal{P}$, and $\mathrm{Exec}_{\mathrm{base}}$ corresponds to standard execution for tools that do not induce page transitions. Here, $a_t$ denotes the tool selected by the agent at outer-loop step $t$, and $\mathcal{T}_{\text{page}}$ denotes the subset of tools that initiate a transition into a new page, which in our design includes \texttt{visit} and \texttt{click}.

As a result, the outer loop receives a compact and goal-aligned\footnote{For this reason, our browser tools that introduce new pages explicitly take the goal as an input parameter.} response $r_t = \mathcal{W}^\star$ instead of the raw page content $\mathcal{P}$, reflecting the core design of the \textit{Nested Browser-Use Framework} and enabling controlled information injection for subsequent reasoning and action. During execution, the agent’s outer-loop is serialized in a structured format: free-form reasoning is enclosed within \texttt{<think>} and \texttt{</think>} tags, tool invocations are wrapped by \texttt{<tool\_call>} and \texttt{</tool\_call>}, and tool responses are encapsulated using \texttt{<tool\_response>} and \texttt{</tool\_response>}. For the inner loop, the temporary workspace is enclosed within \texttt{<useful\_info>} and \texttt{</useful\_info>} tags and returned as the interface to the outer loop.

\subsection{Multi-Task Imitation Learning} \label{sec:learning}

\paragraph{Task and Supervision Sources.}
Comprehensive browser-use capabilities are best elicited under challenging IS tasks. Simple single-hop or multi-hop QA training sets \citep{ yang2018hotpotqa, ho2020constructingmultihopqadataset, mmlu} are insufficient for this purpose, as they can often be solved with only one or a few search queries, without requiring substantive browsing or reasoning.  Accordingly, we adopt SailorFog-QA-V2 \citep{li2025websailorv2bridgingchasmproprietary} as our primary training task, a high-quality QA set designed to elicit complex browsing behaviors that require both effective tool use and multi-step reasoning from IS agents.

\paragraph{Quality Filtering via Rejection Sampling.}
We construct trajectories following the procedure described in \S\ref{sec:framework}. 
Because interleaving reasoning with browser interaction constitutes a challenging agentic capability, and model rollouts do not always exhibit the desired behavior. To encourage correct agentic browsing, we apply rejection sampling \citep{wu2025webdancerautonomousinformationseeking} to the generated trajectories, retaining only high-quality trajectories that satisfy predefined criteria while discarding those that exhibit incorrect or undesirable patterns. This filtering biases supervision toward effective agentic behaviors, avoiding the internalization of spurious patterns. 
Specifically, we apply rejection based on the following three criteria:

\noindent
$\circ$ \textit{\textbf{Format violations.} }Trajectories that do not adhere to the required output format are discarded, such as failing to enclose reasoning content within \texttt{<think>} and \texttt{</think>} tags.

\noindent
$\circ$ \textit{\textbf{Tool-call hallucinations.}} We reject trajectories that contain invalid tool names or tool arguments that cannot be correctly executed.

\noindent
$\circ$ \textit{\textbf{Incorrect final answers.}} We assume that only trajectories leading to a correct final answer provide meaningful supervision for learning agentic browsing and reasoning behaviors, and thus discard trajectories with incorrect outcomes.

Notably, we do not apply additional rejection criteria to intermediate reasoning or browsing steps. This choice preserves diversity in the supervision signal and avoids over-constraining agent behavior with brittle, manually specified rules.

\paragraph{Multi-Task Imitation Learning Objective.}

Because the proposed \textit{Nested Browser-Use Framework} naturally induces multiple learning objectives, training an IS agent requires a multi-task learning formulation. Specifically, the outer loop aims to derive the final answer through interleaved reasoning and tool use, while the inner loop focuses on extracting goal-relevant evidence from pages and incrementally maintaining a temporary workspace.

We first define an outer-loop imitation objective over accepted agent trajectories constructed following \S\ref{sec:framework}. Each trajectory consists of a sequence of outer-loop steps indexed by $t$, where the agent observes a context $c_t$ and produces a serialized continuation that includes reasoning, tool invocations, and tool responses. We optimize a token-level negative log-likelihood objective:
\begin{equation}\label{eq:outer_loss}
\mathcal{L}_{\mathrm{out}}(\theta)
=
\mathbb{E}\left[
\sum_t \sum_j -\log p_\theta\bigl(y_{t,j}\mid c_t, y_{t,<j}\bigr)
\right],
\end{equation}
where $p_\theta$ denotes the parameterized IS agent model and $y_{t,j}$ denotes the $j$-th token in the target serialized IS agent output at step $t$ \citep{chen2023fireact}.

For the inner loop, we supervise the agent model to generate goal-relevant evidence extracted from each page segment and written into the temporary workspace. Let $\mathcal{P}_i$ denote the $i$-th segment of a page and $g_t$ the goal passed from the outer loop. The inner-loop objective is defined as
\begin{equation}\label{eq:inner_loss}
\mathcal{L}_{\mathrm{in}}(\theta)
=
\mathbb{E}\left[
\sum_{t,i}\sum_{j}
-\log p_\theta\bigl(u_{t,i,j}\mid \mathbf{h}_{t,i,j}\bigr)
\right],
\end{equation}
where $\mathbf{h}_{t,i,j}$ denotes the context consisting of the goal $g_t$, page segment $\mathcal{P}_i$, and previously generated tokens $u_{t,i,<j}$, and $u_{t,i,j}$ denotes the $j$-th token of the extracted goal-relevant content for segment $\mathcal{P}_i$.

We jointly optimize the two objectives using a weighted multi-task loss:
\begin{equation}\label{eq:mtl_obj}
\mathcal{L}_{\mathrm{MT}}(\theta)
=
\lambda_{\mathrm{out}}\,\mathcal{L}_{\mathrm{out}}(\theta)
+
\lambda_{\mathrm{in}}\,\mathcal{L}_{\mathrm{in}}(\theta),
\end{equation}
where $\lambda_{\mathrm{out}}$ and $\lambda_{\mathrm{in}}$ balance trajectory-level supervision between outer-loop agentic reasoning with tool use and inner-loop goal-driven page exploration and evidence extraction\footnote{By default, we use equal weighting with $\lambda_{\mathrm{out}} = \lambda_{\mathrm{in}} = 1$.}. Through this multi-task objective, nested browser-use capabilities are jointly trained within a single IS agent model.

\section{Experiments}

\subsection{Setup}

\paragraph{Benchmarks.} We evaluate the proposed NestBrowse on a set of widely recognized and highly challenging deep IS benchmarks. Specifically, we consider English benchmarks including BrowseComp \citep{bc_en} and GAIA \citep{mialon2023gaia}, as well as Chinese benchmarks including BrowseComp-zh \citep{bc_zh} and XBench-DeepSearch (XBench) \citep{xbench}. All benchmarks are web-based QA tasks that require strong agentic reasoning with tool use to locate, navigate, and synthesize hard-to-find web information that is not directly retrievable via simple search queries. 

Following prior work \citep{Li2025webthinker}, we use the 103-question text-only subset of GAIA, while employing the full datasets for the remaining benchmarks. Performance is measured by final answer accuracy. For answer verification, we adopt an \textit{LLM-as-a-Judge} protocol \citep{llmasajudge} using GPT-4.1 \citep{gpt4.1}, following the official evaluation prompts provided by each benchmark.

\paragraph{Baselines.} We comprehensively compare the proposed NestBrowse with open-source IS agents and additionally include several proprietary IS agents for reference. Baseline details are provided in Table~\ref{tab:main_result}, with some results taken from prior work or official leaderboards.

\paragraph{NestBrowse Implementation.} We select two open-source models as base models for training our browser-use IS agents: Qwen3-4B-Thinking-2507 and Qwen3-30B-A3B-Thinking-2507 \citep{yang2025qwen3}. For both models, we set the maximum context length to $128\mathrm{K}$ tokens and cap the number of tool invocations at 100. If an agent fails to produce a final answer upon reaching this limit, the episode is forcibly terminated. 
Following the proposed multi-task training procedure (\S\ref{sec:learning}), we train the 4B model for approximately 1,344 GPU hours and the 30B-A3B model for approximately 4,096 GPU hours on an NVIDIA H20 GPU cluster, resulting in our final models, NestBrowse-4B and NestBrowse-30B-A3B.

\subsection{Overall Performance}

\begin{table*}[ht]
\small
\centering
\caption{Main results on four challenging IS benchmarks. In the Web Toolkit column, \texttt{browser (text)} indicates that only textual elements returned by the browser are used. All resumlts are reported using the \texttt{pass@1} metric. For Xbench, $\dag$ denotes the more challenging \texttt{2510} version; otherwise the \texttt{2505} version.}
\label{tab:main_result}
\resizebox{\textwidth}{!}{
\setlength{\tabcolsep}{6pt} 
\renewcommand{\arraystretch}{1.2} 
\begin{tabular}{@{}l|c|c|c|c|c@{}}
\toprule
\textbf{Model / Framework} & \textbf{Web Toolkit} & \textbf{BrowseComp} & \textbf{BrowseComp-zh} & \textbf{GAIA} & \textbf{XBench} \\
\midrule
\multicolumn{6}{c}{\cellcolor{blue!20}\textit{\textbf{Closed-Source Information-Seeking Agents}}} \\
\midrule
Claude-4-Sonnet \citep{claude4} & \texttt{not reported} & 12.2 & 29.1 & 68.3 & 64.6 \\
Claude-4-Opus \citep{claude4} & \texttt{not reported} & 18.8 & 37.4 & -- & -- \\
Kimi Researcher \citep{kimi_researcher} & \texttt{browser (text)} & -- & -- & -- & 69.0 \\
OpenAI-o4-mini \citep{o3} & \texttt{browser (text)} & 28.3 & 44.3 & -- & -- \\
OpenAI-o3 \citep{o3} & \texttt{browser} & 49.7 & 58.1 & 70.5 & 66.7 \\
OpenAI DeepResearch \citep{openaidr} & \texttt{browser} & 51.5 & 42.9 & 67.4 & -- \\
UI-TARS-2 \citep{wang2025ui} & \texttt{browser} & 29.6 & 50.5 & -- & -- \\
\midrule
\multicolumn{6}{c}{\cellcolor{blue!30}\textit{\textbf{Open-Source Information-Seeking Agents}}} \\
\midrule
ASearcher-Web-32B \citep{asearcher} & \texttt{search, visit} & 5.2 & 15.6 & 52.8 & 42.1 \\
DeepDive-32B \citep{lu2025deepdive} & \texttt{search, visit} & 14.8 & 25.6  & -- & 50.5 \\
DeepDiver-V2-38B \citep{deepdiver-v2} & \texttt{search} & 13.4 & 34.6 & -- & 53.0 \\
Kimi-K2-Instruct-1T \citep{kimi-k2} & \texttt{search, visit} & 14.1 & 28.8 & 57.7 & 50.0 \\
GLM-4.5-355B \citep{zeng2025glm} & \texttt{not reported} & \underline{26.4} & \underline{37.5} & 66.0 & 70.0 \\
WebExplorer-8B \citep{liu2025webexplorer} & \texttt{search, visit} & 15.7 & 32.0 & 50.0 & 53.7 \\
WebDancer-QwQ-32B \citep{wu2025webdancerautonomousinformationseeking} & \texttt{search, visit} & 3.8 & 18.0 & 51.5 & 38.3 \\
WebSailor-32B \citep{li2025websailornavigatingsuperhumanreasoning} & \texttt{search, visit} & 10.5 & 25.5 & 53.2 & 53.3 \\
WebSailor-72B \citep{li2025websailornavigatingsuperhumanreasoning} & \texttt{search, visit} & 12.0 & 30.1 & 55.4 & 55.0 \\
WebShaper-QwQ-32B \citep{tao2025webshaper} & \texttt{search, visit} & -- & -- & 53.3 & 35.0 \\
MiroThinker-32B-DPO-V0.2 \citep{miromind2025mirothinker} & \texttt{search, visit} & 13.0 & 17.0 & 64.1 & -- \\
WebSailor-V2-30B-A3B-SFT \citep{li2025websailorv2bridgingchasmproprietary} & \texttt{search, visit} & 24.4 & 28.3 & 66.0 & 61.7 \\
WebLeaper-30B-A3B-RU \citep{tao2025webleaper} & \texttt{search, visit} & 23.0 & -- & 67.0 & 66.0 \\
\midrule
\textbf{NestBrowse-4B} & \texttt{browser (text)} & 22.4 & 28.4 & \underline{68.9} & \underline{74.0} | \underline{38.0} $\dag$ \\
\textbf{NestBrowse-30B-A3B} & \texttt{browser (text)} & \textbf{31.6} & \textbf{42.6} & \textbf{75.7} & \textbf{75.0} | \textbf{45.0} $\dag$ \\
\bottomrule
\end{tabular}
}
\end{table*}

We evaluate NestBrowse against a broad set of widely adopted IS agent baselines across four challenging deep IS benchmarks. 
As shown in Table~\ref{tab:main_result}, NestBrowse-30B-A3B consistently delivers strong performance, exceeding that of leading open-source IS agents and remaining competitive with, or outperforming, several proprietary systems. Although NestBrowse is trained solely on English data, its browser-use and IS capabilities generalize well to out-of-distribution Chinese benchmarks.

Notably, even NestBrowse-4B achieves competitive performance, exceeding many IS agents with substantially larger parameter counts.
This observation underscores a key insight for IS tasks: \textit{performance is not solely determined by model scale, but is critically influenced by how agents are designed to access, organize, and interact with external information sources.}
From this perspective, we argue that training agentic capabilities on relatively small models is highly valuable~\citep{tongyidr,belcak2025small}.
With appropriate tool abstractions and interaction strategies, small agent models can achieve performance comparable to, or even exceeding, that of substantially larger systems.

\subsection{Analysis of Browser-Use Strategies}

\begin{wraptable}{r}{8cm}
\small
\centering
\caption{Ablation study on browser-use settings. \textit{Simp.} indicates whether the browser toolkit is simplified, and \textit{Extr.} indicates whether goal-relevant content extraction is applied to page responses.
}
\renewcommand{\arraystretch}{1.1} 
\begin{adjustbox}{width=1.0\linewidth}
\centering
\label{tab:ablation}
\begin{tabular}{l|cc|cc}
\toprule
\textbf{Setting} & \textbf{\textit{Simp.}} & \textbf{\textit{Extr.}} & \textbf{GAIA} & \textbf{XBench} \\
\midrule
Naive & \xmark & \xmark & 46.6 & 40.0 \\
Simplified & \cmark & \xmark & 55.3 & 40.0 \\
Compressed & \xmark & \cmark & 60.2 & 61.0 \\
\midrule
\textbf{NestBrowse} & \cmark & \cmark & \textbf{73.8} & \textbf{71.0} \\
\bottomrule
\end{tabular}
\end{adjustbox}
\end{wraptable}

To validate the design choices of NestBrowse, we conduct an ablation study (Table~\ref{tab:ablation}) comparing three browser-use strategies. \textit{Naive} uses a standard browser-use setup without tool simplification or page-level information extraction; \textit{Simplified} adopts the four-action toolkit of NestBrowse without goal-conditioned extraction; and \textit{Compressed} applies goal-relevant page content extraction while retaining the original toolkit.

To isolate the impact of strategy design, all variants are executed using the same strong agent model, GPT-OSS-120B \citep{gptoss}. Both toolkit simplification and goal-relevant page content extraction independently improve performance, with extraction yielding larger gains by reducing redundancy and focusing agent reasoning on relevant information. Combining both components, NestBrowse achieves the strongest performance, exhibiting a clear additive effect.

\subsection{Context Efficiency of NestBrowse}

\begin{wrapfigure}{r}{0.6\linewidth}
    \centering
    \includegraphics[width=1\linewidth]{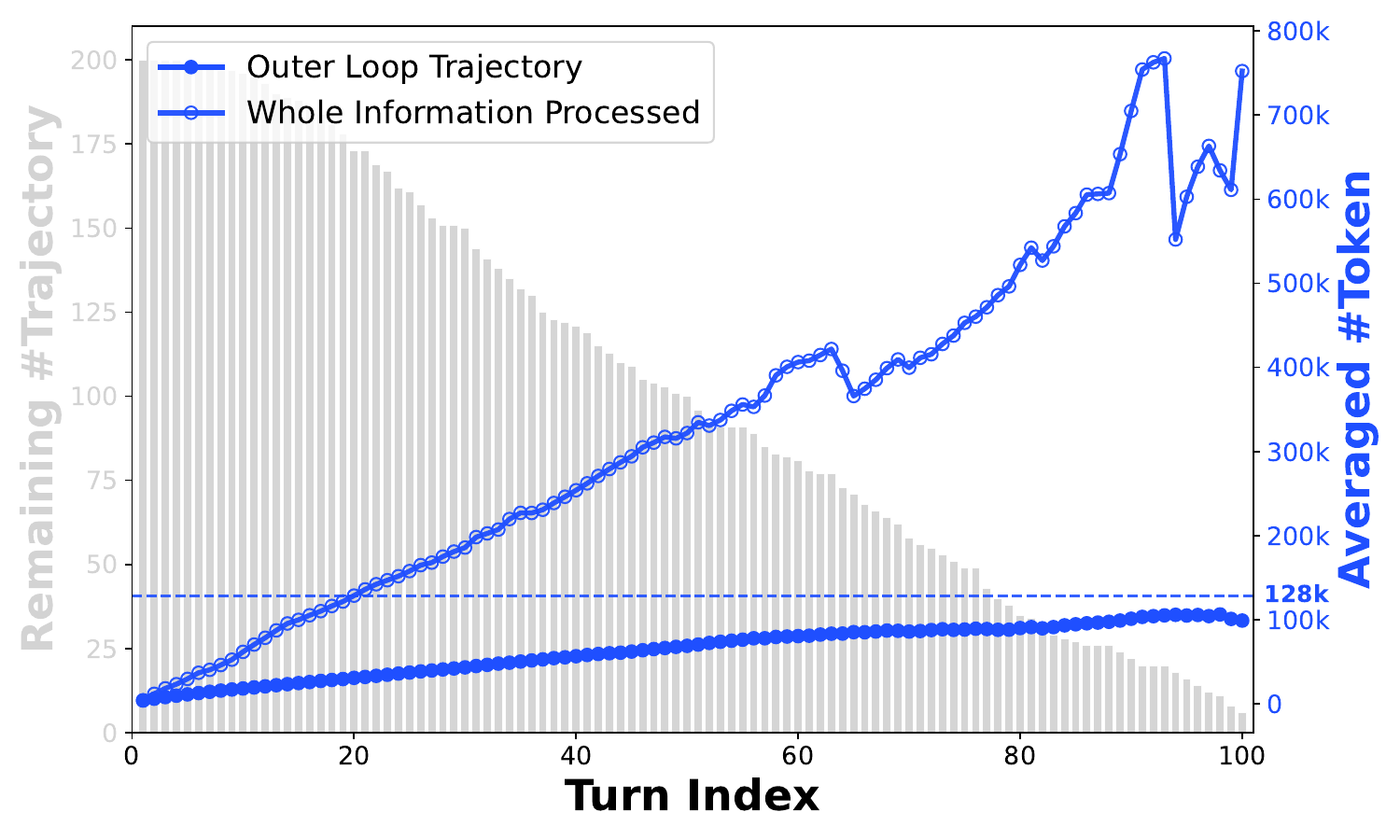}
    \caption{Average context length over tool-call turns for NestBrowse-30B-A3B on the BrowseComp subset. Gray bars denote the number of remaining active trajectories at each turn.}
    \label{fig:context_efficiency}
\end{wrapfigure}

We analyze the context efficiency of NestBrowse to assess its ability to maintain controlled information flow during browser-use. As shown in Figure~\ref{fig:context_efficiency}, \textit{Whole Information Processed} denotes the total page content handled jointly by the outer and inner loops, while the other curve shows the context length maintained in the outer (agentic reasoning) loop. After approximately \textbf{20} tool-call turns, the total processed information already exceeds the agent’s maximum context limit ($128\mathrm{K}$ tokens). Without NestBrowse, execution would terminate at this point, even though about \textbf{85\%} of tasks remain unfinished. By injecting only goal-relevant content into the outer loop, NestBrowse keeps the agent context within feasible limits throughout execution, substantially improving the practicality and effectiveness of browser-based information seeking.

\begin{wrapfigure}{r}{0.55\linewidth}
    \centering
    \includegraphics[width=1\linewidth]{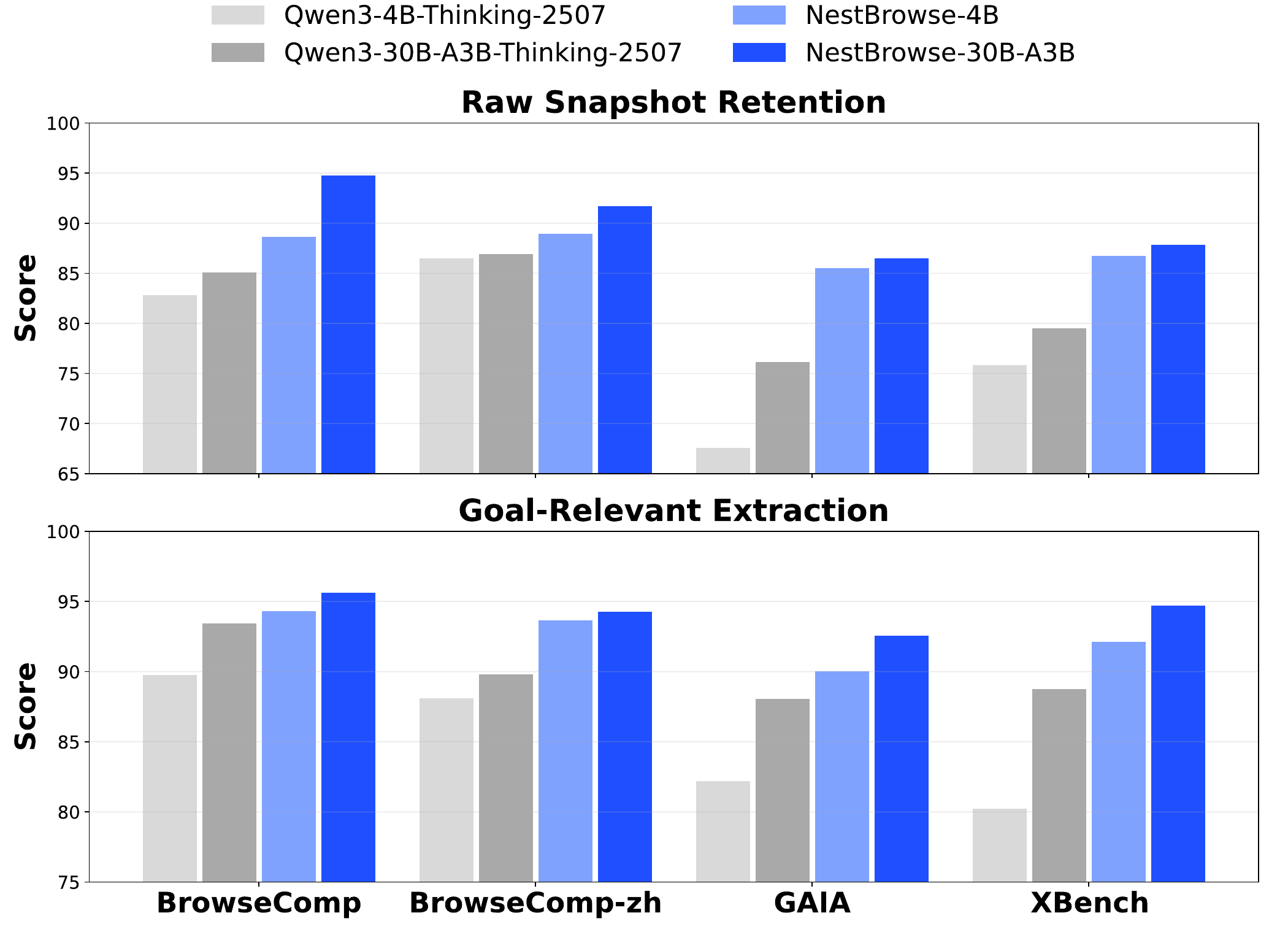}
    \caption{Intra-page exploration performance over inner-loop instances from 100 trajectories per benchmark.}
    \label{fig:page_reading_acc}
\end{wrapfigure}

\subsection{Analysis of Intra-Page Exploration}

We further analyze the effectiveness of intra-page exploration in the NestBrowse inner loop, and evaluate the model along two dimensions using GPT-4.1 as the judge model: \textit{raw snapshot retention}, which measures how well the agent preserves original page snapshot to support subsequent 
interactions, and \textit{goal-relevant extraction accuracy}, 
which assesses whether the extracted content aligns with the given goal. As shown in Figure~\ref{fig:page_reading_acc}, compared to the base agent model, NestBrowse achieves substantial improvements on both metrics. These results validate the effectiveness of our multi-task training scheme, which simultaneously strengthens outer-loop agentic reasoning and inner-loop intra-page exploration, integrating complex browser-use capabilities into a single IS agent model.

\begin{wraptable}{l}{10cm}
\small
\centering
\caption{Deep IS performance with different inner-loop models on the BrowseComp subset (100 examples).}
\renewcommand{\arraystretch}{1.1} 
\begin{adjustbox}{width=1.0\linewidth}
\centering
\label{tab:page_reading_overall_effect}
\begin{tabular}{l|l|c}
\toprule
\textbf{Outer-Loop Model} & \textbf{Inner-Loop Model} & \textbf{BrowseComp} \\
\midrule
\multirow{3.2}{*}{NestBrowse-30B-A3B} & NestBrowse-4B & 24.0 \\
& NestBrowse-30B-A3B & 35.0 \\
& GPT-OSS-120B & 36.0 \\
\bottomrule
\end{tabular}
\end{adjustbox}
\end{wraptable}

We also analyze the impact of inner-loop intra-page exploration quality on outer-loop agentic reasoning. As shown in Table~\ref{tab:page_reading_overall_effect}, weaker inner-loop models degrade outer-loop performance, while stronger ones yield corresponding improvements, indicating a clear positive relationship between the two. This finding underscores intra-page exploration as a critical component of information acquisition in IS tasks and provides empirical justification for jointly training the outer and inner loops under a unified multi-task objective.

\subsection{Case Study: Beyond Static Web Access}

\begin{figure*}[htbp]
    \centering
    \includegraphics[width=1\textwidth]{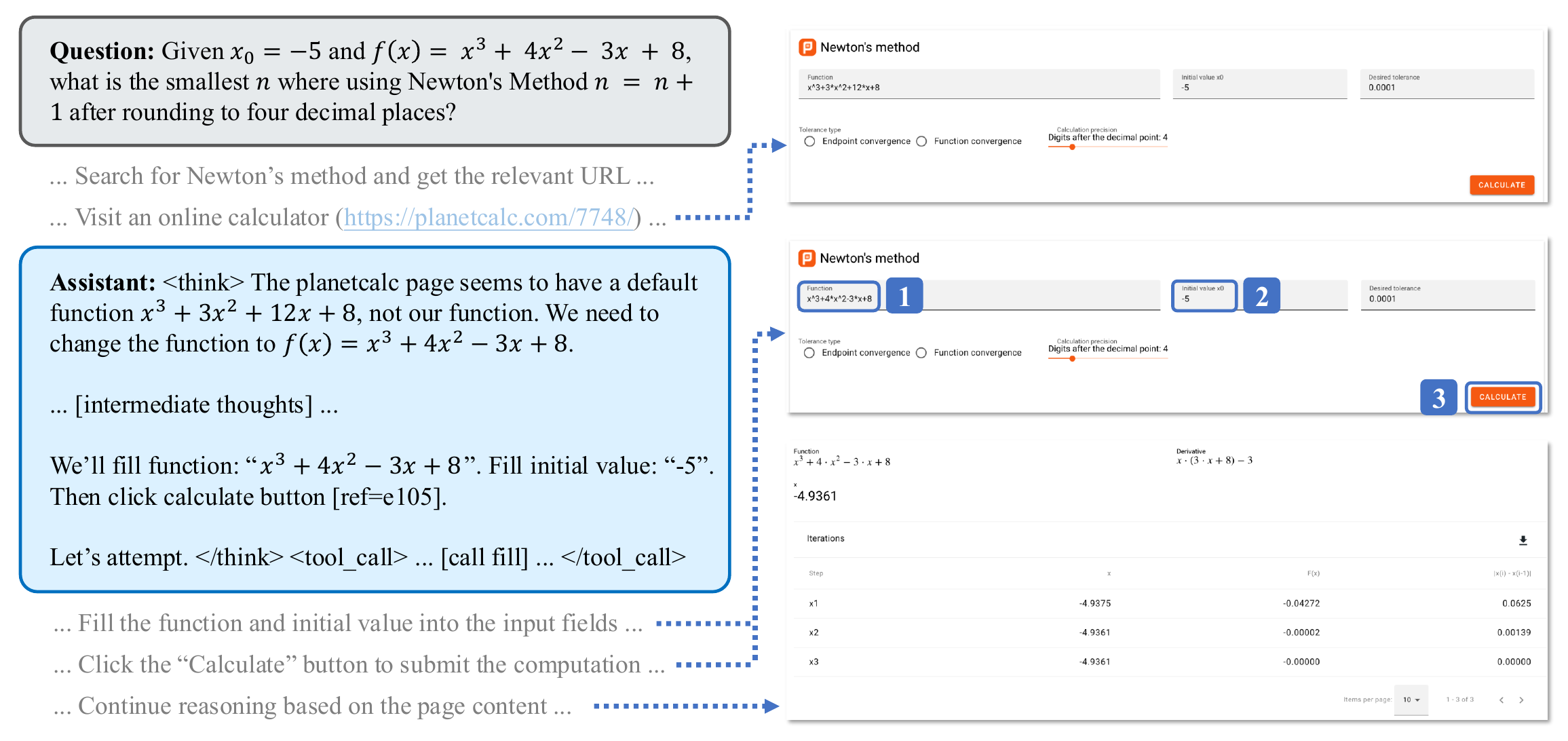}
    \caption{A case study where an online browser-based calculator is used for numerical computation.}
    \label{fig:case_study}
\end{figure*}

We further analyze NestBrowse through a representative case study, shown in Figure~\ref{fig:case_study}. 
When solving a GAIA task involving numerical computation, NestBrowse-30B-A3B adopts a browser-based strategy by invoking an online calculator to perform the required computation. 
For a small-scale model, such numerical reasoning would be difficult to handle reliably through internal reasoning alone. 
By leveraging browser-use to exploit a computational utility embedded within the webpage, the agent substantially reduces its reasoning burden and successfully completes the task.

This case highlights a key advantage of browser-use beyond retrieving static page content. Rather than serving solely as a medium for information access, browser-use enables agents to exploit the rich functional ecosystem of the web, including interactive utilities that are inaccessible through static page fetching alone. From this perspective, browser-use can be viewed as a form of \textbf{meta tool-use}, where the effective toolkit available to an agent encompasses the broad collection of tools implicitly embedded within the web itself. NestBrowse further facilitates the acquisition of such meta tool-use capabilities, enabling agents to solve complex IS tasks in a more flexible and effective manner.

\section{Conclusion}

We present Nested Browser-Use Learning (\textbf{NestBrowse}), a learning framework that equips information-seeking (IS) agents with a minimal yet complete browser toolkit and a nested browser-use paradigm for efficient deep web information access. By decoupling agentic reasoning from intra-page exploration and training these capabilities jointly, NestBrowse enables strong deep IS performance across relatively small model scales. Our results suggest that principled browser-use abstraction and interaction strategies are key to solving complex IS tasks, even without relying on large-scale agent models.
\section{Limitations and Future Work}

While our proposed NestBrowse successfully enables small models to acquire complex browser-use capabilities and interact with real browsers to collect information for downstream reasoning, we intentionally restrict our browser modeling to textual content. This design choice is motivated by the need to isolate the effects of tool abstraction and interaction paradigms, and to maintain a controlled and interpretable experimental setting. Incorporating additional modalities such as vision or audio would introduce substantial modeling and system complexity, making it difficult to disentangle the contributions of browser-use learning from those of multimodal perception. Nevertheless, non-textual modalities often convey important information in real-world browsing scenarios. Extending the proposed NestBrowse to support multimodal information acquisition and reasoning represents a promising future direction. 

\clearpage


\bibliography{custom}
\bibliographystyle{colm2024_conference}

\clearpage
\appendix

\section{Appendix}
\label{sec:appendix}

\subsection{Implementation Details: Prompts}

We report all prompts used in the NestBrowse. We begin with the prompts for outer-loop agentic reasoning. Following the standard ReAct-style function-calling format, we adopt the system prompt shown below. The user prompt in the outer loop corresponds directly to the user query, without any additional instructional or guiding content.

\begin{tcolorbox}[
    enhanced,
    breakable,
    sharp corners,
    boxrule=0pt,
    leftrule=3pt,              
    colback=gray!5,            
    colframe=gray!80!black,    
    fonttitle=\bfseries\color{gray!80!black},
    title=PROMPT,              
    detach title,              
    before upper={\textcolor{gray!50!black}{\textbf{SYSTEM PROMPT (OUTER LOOP)}} \vspace{0.2cm}\\}
]
You are a browser-use agent. Your core function is to conduct thorough, multi-source investigations into any topic. You must handle both broad, open-domain inquiries and queries within specialized academic fields. For every request, synthesize information from credible, diverse sources to deliver a comprehensive, accurate, and objective response. When you have gathered sufficient information and are ready to provide the definitive response, you must enclose the entire final answer within <answer></answer> tags.
\\\\
\# Tools
\\\\
You may call one or more functions to assist with the user query.
\\\\
You are provided with function signatures within <tools></tools> XML tags:\\
<tools>\\
\{BROSWER\_TOOLS\_SCHEMA\}\\
</tools>
\\\\
For each function call, return a json object with function name and arguments within <tool\_call></tool\_call> XML tags:\\
<tool\_call>\\
\{"name": <function-name>, "arguments": <args-json-object>\}\\
</tool\_call>
\end{tcolorbox}

We then present the prompts used for inner-loop page exploration. This component is responsible for extracting goal-relevant content from a webpage given a specified goal, and producing outputs in a predefined JSON format. The corresponding system prompt is shown below. The content enclosed within the \texttt{<useful\_info>} and \texttt{</useful\_info>} tags corresponds to the temporary workspace described in \S\ref{sec:framework}, which incrementally maintains the extracted goal-relevant information during intra-page exploration.

\begin{tcolorbox}[
    enhanced,
    breakable,
    sharp corners,
    boxrule=0pt,
    leftrule=3pt,              
    colback=gray!5,            
    colframe=gray!80!black,    
    fonttitle=\bfseries\color{gray!80!black},
    title=PROMPT,              
    detach title,              
    before upper={\textcolor{gray!50!black}{\textbf{SYSTEM PROMPT (INNER LOOP)}} \vspace{0.2cm}\\}
]
You must answer only by outputting a single valid JSON object, with no extra text before or after it. 
\\\\
Your task: given webpage content and a user goal, extract and organize the useful information according to the following schema: \{"rational": "string", "evidence": "string", "summary": "string"\}. 
\\\\
Follow these rules for each field: \\
1) rational: Locate the **specific sections/data** directly related to the user's goal within the webpage content. \\
2) evidence: Identify and extract the **most relevant information** from the content, never miss any important information, output the **full original context** of the content as far as possible, it can be more than three paragraphs. \\
3) summary: Organize into a concise paragraph with logical flow, prioritizing clarity and judge the contribution of the information to the goal. 
\\\\
Formatting requirements: Output only one valid JSON object wrapped inside <useful\_info> and </useful\_info> tags: use double quotes (") for all keys and string values, no trailing commas, and the top-level structure must be exactly: \{"rational": "...", "evidence": "...", "summary": "..."\}.
\end{tcolorbox}

For the inner loop, the user prompt combines the given webpage content with the specified goal, instructing the model to extract relevant information and produce outputs in the required format.

\begin{tcolorbox}[
    enhanced,
    breakable,
    sharp corners,
    boxrule=0pt,
    leftrule=3pt,              
    colback=gray!5,            
    colframe=gray!80!black,    
    fonttitle=\bfseries\color{gray!80!black},
    title=PROMPT,              
    detach title,              
    before upper={\textcolor{gray!50!black}{\textbf{USER PROMPT (INNER LOOP)}} \vspace{0.2cm}\\}
]
Please process the following webpage content and user goal to extract relevant information:
\\\\
\#\# **Webpage Content** \\
\{raw\_response\}
\\\\
\#\# **User Goal**\\
\{goal\}
\\\\
\#\# **Task Guidelines**\\
1. **Content Scanning for Rational**: Locate the **specific sections/data** directly related to the user's goal within the webpage content.\\
2. **Key Extraction for Evidence**: Identify and extract the **most relevant information** from the content, you never miss any important information, output the **full original context** of the content as far as possible, it can be more than three paragraphs.\\
3. **Summary Output for Summary**: Organize into a concise paragraph with logical flow, prioritizing clarity and judge the contribution of the information to the goal.
\\\\
**Final Output Format using JSON format has "rational", "evidence", "summary" feilds**
\end{tcolorbox}

When a webpage is partitioned into multiple segments, the inner loop iteratively updates the workspace. Specifically, the \texttt{evidence} and \texttt{summary} fields from the workspace produced in the previous iteration are extracted and appended to the current user prompt (incremental). This provides the model with the information already collected, preventing redundant extraction and enabling incremental workspace maintenance. Upon completion of the inner loop, the final workspace is returned to the outer loop for subsequent reasoning.

\begin{tcolorbox}[
    enhanced,
    breakable,
    sharp corners,
    boxrule=0pt,
    leftrule=3pt,              
    colback=gray!5,            
    colframe=gray!80!black,    
    fonttitle=\bfseries\color{gray!80!black},
    title=PROMPT,              
    detach title,              
    before upper={\textcolor{gray!50!black}{\textbf{USER PROMPT INCREMENTAL (INNER LOOP)}} \vspace{0.2cm}\\}
]
Please process the following webpage content and user goal to increamentally extract relevant information:
\\\\
\#\# **Webpage Content** \\
\{raw\_response\}
\\\\
\#\# **User Goal**\\
\{goal\}
\\\\
\#\# **Task Guidelines**\\
1. **Content Scanning for Rational**: Locate the **specific sections/data** directly related to the user's goal within the webpage content\\
2. **Key Extraction for Evidence**: Identify and extract the **most relevant information** from the content, you never miss any important information, output the **full original context** of the content as far as possible, it can be more than three paragraphs.\\
3. **Summary Output for Summary**: Organize into a concise paragraph with logical flow, prioritizing clarity and judge the contribution of the information to the goal.
\\\\
\#\# **Existing Evidence**\\
\{existing\_evidence\}

\#\# **Existing Summary**\\
\{existing\_summary\}
\\\\
Note: Existing extracted evidence and summaries are already provided. You must build upon and integrate these existing pieces of information to perform incremental processing. Produce a consolidated final result that incorporates both the provided and newly added information, without indicating which parts are new or incremental.
\\\\
**Final Output Format using JSON format has "rational", "evidence", "summary" feilds**
\end{tcolorbox}

\end{document}